\documentclass{INTERSPEECH2023}

\interspeechcameraready

\usepackage{float}
\usepackage{listings}
\usepackage{xcolor}
\definecolor{codegreen}{rgb}{0,0.6,0}
\definecolor{codegray}{rgb}{0.5,0.5,0.5}
\definecolor{codepurple}{rgb}{0.58,0,0.82}
\definecolor{backcolour}{rgb}{0.95,0.95,0.92}
\lstdefinestyle{mystyle}{
    backgroundcolor=\color{backcolour},   
    commentstyle=\color{codegreen},
    keywordstyle=\color{magenta},
    numberstyle=\tiny\color{codegray},
    stringstyle=\color{codepurple},
    basicstyle=\ttfamily\footnotesize,
    breakatwhitespace=false,         
    breaklines=true,                 
    captionpos=b,                    
    keepspaces=true,                 
    numbers=left,                    
    numbersep=5pt,                  
    showspaces=false,                
    showstringspaces=false,
    showtabs=false,                  
    tabsize=2
}

\title{Generative Adversarial Training for Text-to-Speech Synthesis Based on Raw Phonetic Input and Explicit Prosody Modelling}
\name{Tiberiu Boros$^1$, Stefan Daniel Dumitrescu$^1$, Ionut Mironica$^1$, Radu Chivereanu$^1$}
\address{
  $^1$Adobe Systems, Romania}
\email{boros@adobe.com, sdumitre@adobe.com, mironica@adobe.com, rchivereanu@adobe.com}

\begin{document}

\maketitle
 
\begin{abstract}
We describe an end-to-end speech synthesis system that uses generative adversarial training. We train our Vocoder for raw phoneme-to-audio conversion, using explicit phonetic, pitch and duration modeling. We experiment with several pre-trained models for contextualized and decontextualized word embeddings and we introduce a new method for highly expressive character voice matching, based on discreet style tokens.
\end{abstract}

\section{Introduction}

Typically, the architecture of text-to-speech (TTS) synthesis systems is driven by two meta-challenges: (a) take a symbolic representation (usually raw text), process it and get to an intermediate representation of speech that encompasses the message and inferred prosodic effects such as word emphasis, emotion etc. - this is achieved by the \textbf{text-processing backend} and (b) generate high quality audio from this intermediate representation - this is achieved by the \textbf{vocoder}.

Usually the first step involves phonetic transcription, which is somewhat straight-forward, followed by the reconstruction of prosody-related information, which translates into \textbf{explicitly or implicitly modeling duration and pitch contours} for the synthesized audio. This is a highly complex task, mainly because the symbolic representation (text) strips away such information. Accurate prediction of pitch and duration requires inferring the speaker's emotional state as well as being able to capture the subtleties of the message being conveyed by the author. This can only be achieved by using both local and global contexts. The local context means everything that can be found withing a limited window of tokens (usually a paragraph) that surround the utterance being synthesized, while global context is an umbrella of metadata regarding the characters (age, gender, speaking style, accent etc.) and any other relevant information that would help infer local emotions that are being expressed in the utterance. This task is sometimes difficult even for humans if they are not professional speakers. 

However, most text-processing backends simply ignore the global context and rely only on information that can locally be extracted from text \cite{zen2007hmm,wang2017tacotron}. Our system follows the same pattern, given that inferring character information for the global context is a highly  difficult task.

Recently, the major leap in text-to-speech synthesis was centered around \textbf{vocoders} \cite{oord2016wavenet}. It introduced a new model called WaveNet for synthesizing speech that immediately achieved remarkable results, far beyond the state-of-the-art at that point. The idea of using a neural network to directly synthesize speech (in the time-domain) without any pre-processing (i.e. historically carried out in the frequency domain) was both bold and novel. While some focused on making WaveNet work in real-time \cite{paine2016fast,oord2018parallel}, others tried to build new neural models for vocoding, such as SampleRNN \cite{mehri2016samplernn} and WaveRNN \cite{kalchbrenner2018efficient}.

The previously mentioned vocoders belong to the class of auto-regressive models, which are inherently slow in the generative process. Trying to move away from these models resulted in flow-based models such as \cite{kim2018flowavenet, peng2020clarinet}, but from our experience, the performance of these models is modest, when compared with their auto-regressive counterparts. Finally, advances in generative adversarial training, yielded a new class of parallel vocoders, such as those presented in \cite{kumar2019melgan, kong2020hifi}, that are part of the current state-of-the-art solutions to vocoding.


Our proposed system is an \textbf{end-to-end speech synthesis model that jointly optimizes portions of the text-processing backend along with the vocoder}. Additionally, we rely on a transformer model (BERT, \cite{devlin2018bert}) to provide \textbf{contextualized embeddings for our backend}. The text transformer is fine-tuned with the entire ensemble. 

In what follows, we will briefly present related work (Section \ref{sect:related_work}), go through our proposed architecture with training details (Section \ref{sect:architecture}), present the evaluation results (Section \ref{sect:results}) and discuss the current state of the system as well as future plans (Section \ref{sect:conclusion}).

\section{Related work}
\label{sect:related_work}

Several end-to-end solutions have been previously proposed, mostly by combining Tacotron-based architectures with a vocoder (e.g. WaveNet, ParallelWavenet, WaveRNN, Clarinet, Flowavenet, MelGAN, HifiGAN, etc.). In most cases, the vocoder was pre-trained to perform Mel-spectrogram inversion and then optimized on the forced-aligned outputs of the text-processing back-end, in order to cope with the artefacts introduced in the synthetic spectrogram. 

Joint end-to-end optimization was proposed in the original WaveNet paper \cite{oord2016wavenet} (where the authors conditioned the model on an intermediate phonetico-prosodic representation) as well as in several other works \cite{sotelo2017char2wav, mehri2016samplernn, lim2022jets, liu2022delightfultts}.

\section{Proposed architecture}
\label{sect:architecture}

\begin{figure*}[h]
  \centering
  \includegraphics[width=0.7\linewidth]{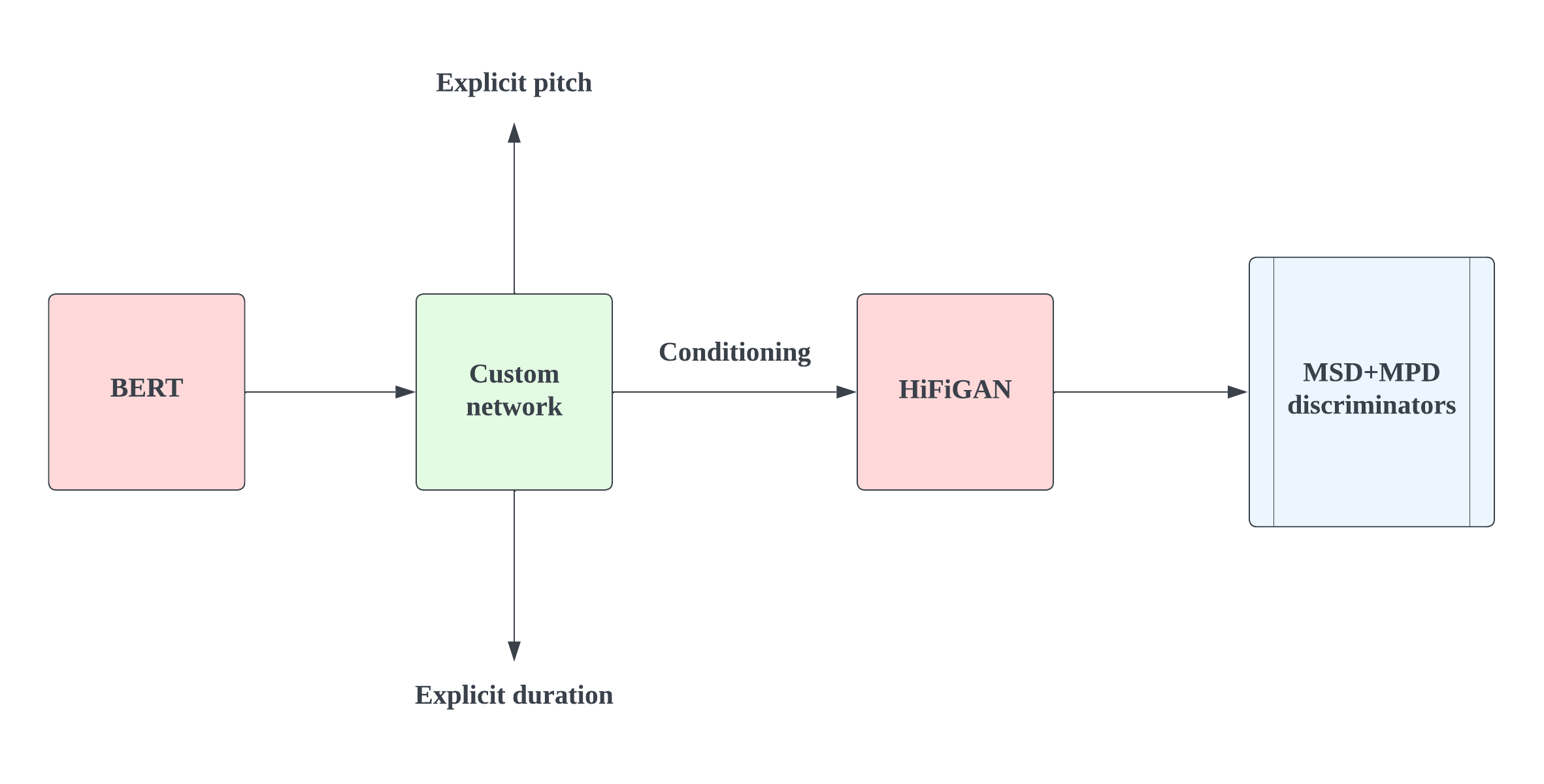}
  \caption{Our end-to-end solution for text-to-speech processing: BERT embeddings, custom network for prosody synthesis and HiFiGAN for vocoding. The system is trained end-to-end, except for the phonemizer which is a different model and it is not illustrated here.}
  \label{fig:architecture-c}
\end{figure*}

Our system is composed of two modules: (a) the main text-to-speech module which \textbf{combines BERT and HiFiGAN with a custom network architecture for prosody prediction} with GAN conditioning embeddings and (b) the phonemizer, which is responsible for generating a hybrid representation of the text, combining phonemes with punctuation (including spaces).

Section \ref{sect:tts} discusses the main TTS portion of our system, Section \ref{sect:phonemizer} introduces the model responsible for phonetic transcription and Section \ref{sect:training} handles training and model selection.

\subsection{Main text-to-speech network architecture}
\label{sect:tts}

Our approach (see Figure \ref{fig:architecture-c}) uses \textbf{contextualized word embeddings} generated by a BERT-based model, specifically Camembert \cite{martin2019camembert}, with a custom network architecture used for modeling prosody (pitch and duration) and for conditioning HiFiGAN. 

\begin{figure*}[h]
  \centering
  \includegraphics[width=0.98\linewidth]{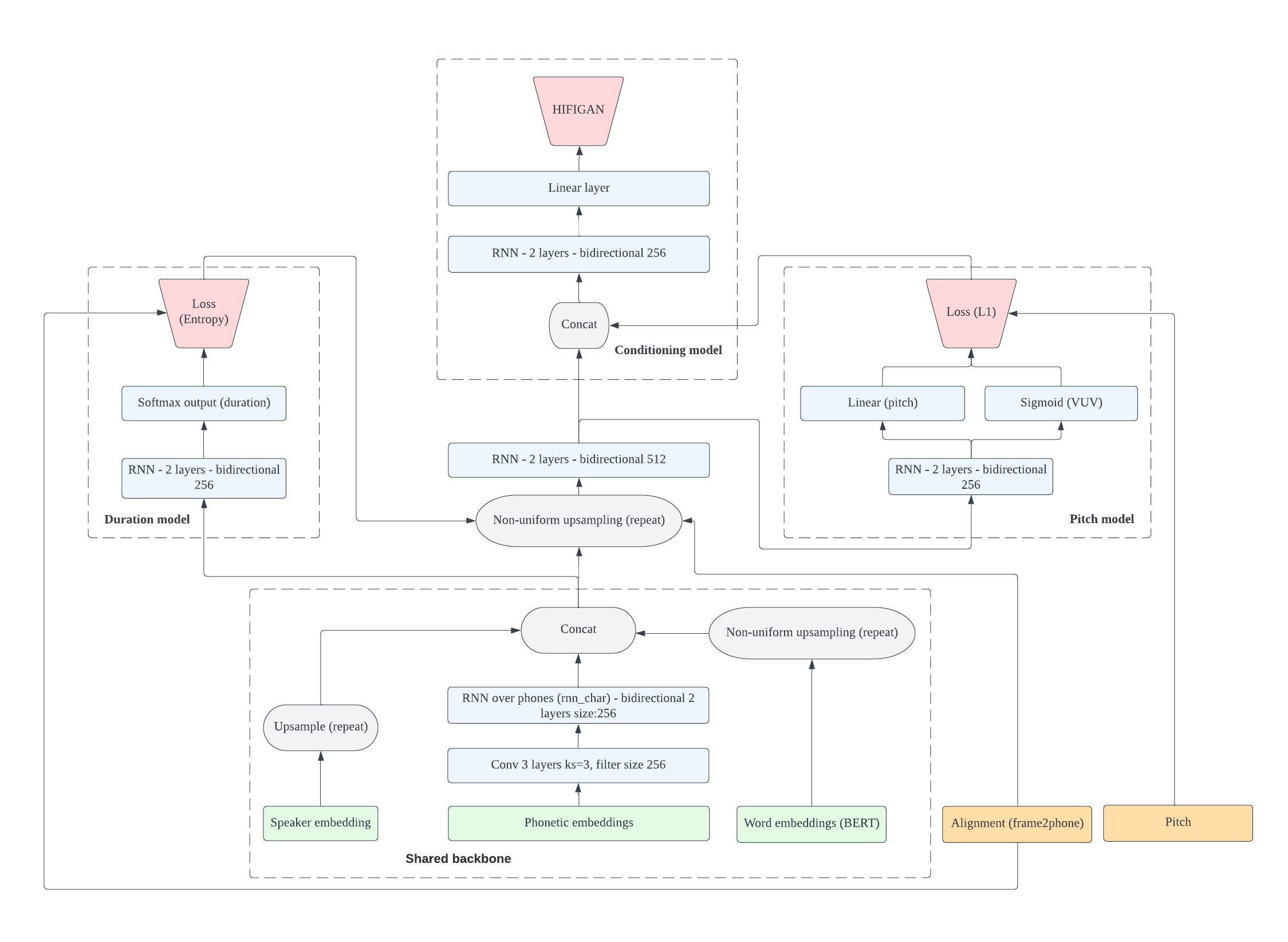}
  \caption{Detailed architecture of the custom network used for synthesis (duration, pitch and HifiGAN conditioning signals)}
  \label{fig:architecture}
\end{figure*}

Both duration and pitch are modeled explicitly: duration as a discreet distribution (softmax), and pitch as a continuous variable combined with a gate for voiced/unvoiced decision (see Figure \ref{fig:architecture}). Training is performed in a forced alignment manner, meaning that we use gold-standard duration to do the non-uniform upsampling of the phonetic representation of the input.
To extract ground truth for pitch training we use
RAPT \cite{rapt}. Additionally, for duration, we use provided annotations.

The custom network architecture is composed of \textbf{three parallel stacks of recurrent neural networks}, more precisely bi-directional Long-Short-Term-Memory-Networks (BiLSTMs). One stack handles duration modeling, the next stack pitch modeling and the final stack is used as input for HiFiGAN. They share a sub-network (backbone) composed of 3 layers of convolutional neural networks and a BiLSTM. The backbone takes as input phonetic embeddings, and the output of the backbone is concatenated with BERT embeddings. For every input phoneme, we align the corresponding output of the backbone with the word embedding produced by BERT, for the word that contained that phoneme. 

The entire system is trained in parallel, but it's not jointly optimized, since we use gold standard data for portions of the training. More spefically, the conditioning for HiFiGAN is acquired through a single stack of RNNs, which receives the forced-aligned input. Furthermore, we simultaneously train the other two stacks of RNNs to predict both the alignments (duration) and pitch. While these modules do share the backbone, it is unlikely that they reach an optimal point at the same time (after the same number of steps). BERT is also optimized with the entire system.

Given that phonemes have different durations, we use non-uniform upsampling, by repeating the same embedding as many times as it takes for the input to match the alignments. For speed, we use PyTorch indexing to carry out the non-uniform upsampling, instead of naively repeating and concatenating vectors.

There are \textbf{two layers of non-uniform upsampling throughout the model}. The first layer refers to upsampling the BERT embeddings to match the corresponding phonemes of each word (token) and the second one refers to upsampling phoneme-level embeddings based on their durations (predicted or gold-standard). Because BERT-based models use either the SentencePiece Tokenizer \cite{kudo2018sentencepiece} or Byte-Pair-Encondings (BPEs) \cite{gage1994bpeorig, sennrich2015bpe} there are major differences between the sub-tokens produced by these tokenizers and the actual words. When a word gets split into multiple sub-tokens by the BERT Tokenizer, we use the contextualized embedding of the first token as the embedding of that word.

\begin{table*}[h]
\caption{The user opinion score obtained by our system, detailing median, mean, mean absolute deviation (MAD) and standard deviation (stdev)}
\label{table:results}
\centering
\begin{tabular}{|lllll|}
\hline
\multicolumn{1}{|l|}{\textbf{Description}}                                                                & \textbf{Median} & \textbf{MAD} & \textbf{Mean} & \textbf{StDev} \\ \hline
\multicolumn{5}{|c|}{\textbf{NEB}}                                                                                                                                          \\ \hline
\multicolumn{1}{|l|}{\textbf{MOS scores - per system for the speech   experts (SE)}}                      & 4.0             & 1.48         & 4.0           & 0.99           \\
\multicolumn{1}{|l|}{\textbf{MOS scores - per system for the paid   listeners (SP)}}                      & 4.0             & 1.50         & 3.6           & 1.06           \\
\multicolumn{1}{|l|}{\textbf{MOS scores - per system for the online   volunteers (SR)}}                   & 4.0             & 0.74         & 4.2           & 0.81           \\
\multicolumn{1}{|l|}{\textbf{MOS scores - per system for the   non-speech experts (N-SE) and native}}     & 4.0             & 1.48         & 3.6           & 1.06           \\
\multicolumn{1}{|l|}{\textbf{MOS scores - per system for the speech   experts (SE) and native}}           & 4.0             & 1.48         & 3.8           & 1.08           \\
\multicolumn{1}{|l|}{\textbf{MOS scores - per system for the   non-speech experts (N-SE) and non-native}} & 4.0             & 0.74         & 4.3           & 0.67           \\
\multicolumn{1}{|l|}{\textbf{MOS scores - per system for the speech   experts (SE) and non-native}}       & 4.0             & 1.48         & 4.3           & 0.78           \\ \hline
\multicolumn{5}{|c|}{\textbf{AD}}                                                                                                                                           \\ \hline
\multicolumn{1}{|l|}{\textbf{MOS scores - per system for the speech   experts (SE)}}                      & 3.0             & 1.48         & 3.3           & 1.00           \\
\multicolumn{1}{|l|}{\textbf{MOS scores - per system for the paid   listeners (SP)}}                      & 4.0             & 1.50         & 3.5           & 1.07           \\
\multicolumn{1}{|l|}{\textbf{MOS scores - per system for the online   volunteers (SR)}}                   & 4.0             & 0.00         & 4.1           & 0.83           \\
\multicolumn{1}{|l|}{\textbf{MOS scores - per system for the   non-speech experts (N-SE) and native}}     & 4.0             & 1.50         & 3.5           & 1.07           \\
\multicolumn{1}{|l|}{\textbf{MOS scores - per system for the speech   experts (SE) and native}}           & 3.0             & 1.48         & 3.2           & 1.06           \\
\multicolumn{1}{|l|}{\textbf{MOS scores - per system for the   non-speech experts (N-SE) and non-native}} & 4.0             & 0.00         & 4.2           & 0.62           \\
\multicolumn{1}{|l|}{\textbf{MOS scores - per system for the speech   experts (SE) and non-native}}       & 3.0             & 0.00         & 3.6           & 0.76           \\ \hline
\end{tabular}
\end{table*}

\subsection{Hybrid grapheme-to-phoneme conversion}
\label{sect:phonemizer}

Normally, grapheme-to-phoneme conversion would require many-to-many alignments between the input text and the output phonetic sequence. Usually, this is achieved by an attention mechanism, relying on the model to learn how to map the input sequence composed of $n$ graphemes to an output sequence composed of $m$ phonemes. However, the dataset we trained the system on already contains alignments between input and output sequences. Furthermore, the mapping is 1:1 and is achieved by using a special token (``\verb"-"'') for inputs that don't map to any phonemes in the output and by concatenating the phonetic representations of the output, whenever one input grapheme corresponds to multiple phonemes. Thus, our grapheme to phoneme module is implemented as a sequence labeling network, where every input has exactly one label. 

The architecture is fairly simple, composed of a stack of three convolutional layers, a stack of BiLSTMs and a softmax output. 

In the original dataset, punctuation was either transcribed using the special void character (``\verb"-"'') or using a special token for silence, whenever the case. We found it better to add punctuation (including spaces) to the output of the phonemizer, since the type of punctuation (``,'', ``.'', ``?'', ``!'', etc.) provides important prosodic cues that influence both duration and pitch.

The system was trained to transcribe entire sentences and have access to the whole context of every word, because the pronunciation of some words is dependent on the context, influenced either by morphology (the case of homographs) or by phonetics (e.g. the pronunciation of the word ``the'' in English is dependent on the first sound of the next word).

\subsection{Training}
\label{sect:training}

The main network architecture was trained with a decaying learning rate of $2 \cdot 10^{-4}$ with a decay rate of $10^{-5}$ per step for 1M steps. We performed no model selection for this network and we just relied on the last version of the model once training was done. The pretrained BERT model (Camembert) was also fine-tuned along our custom network and HiFiGAN, but with a fixed learning rate of $10^{-6}$.

For training the phonemizer, we split the original dataset into a training set (90\% of the data) and a validation set (10\% of the data). After each epoch we evaluated the accuracy of the model and computed the phoneme accuracy rate (PAR) and the sentence accuracy rate (SAR - perfectly translated sentences). We used the sentence accuracy rate for model selection, meaning that in the end we used the model with the highest SAR. For early stopping, we used a patience value of 20, meaning the training was stopped if 20 epochs passed without any improvement on the SAR score.

Some notes about the training and other experiments:

\begin{enumerate}
    \item We experimented with multiple BERT models for French, and, based on our informal listening tests they seemed to generate sub-optimal results compared to CamemBERT;
    \item De-contextualized embeddings based on FastText \cite{bojanowski2017fasttext} are currently supported by our system and were used in our experiments. However, the resulting audio was lacking emotions, character voices and ad-hoc pauses;
    \item The model was trained on an NVIDIA RTX 3090 with 24 GB of RAM, using a batch-size of 16 for approximately 3 weeks;
    \item Not optimizing BERT in an end-to-end manner yielded sub-optimal results;
    \item We tested other configurations of learning rates for BERT, but the $10^{-6}$ seemed to provide better results. However, none of the model configurations were trained for 1M steps to check if this assumption holds;
    \item Conditioning HiFiGAN on the mel spectrogram and training the system as two different modules (one to convert text into a mel-spectrogram and the other to generate audio based on this spectrogram) always resulted in audio artifacts, which were not removed even by fine-tuning the system.
\end{enumerate}

\section{Evaluation}
\label{sect:results}

The evaluation of the system was carried out as part of Blizzard Challenge 2023. The challenge featured a French language dataset \cite{blizzard-dataset} composed of two speakers: Nadine Eckert-Boulet (NEB) and Aurélie Derbier (AD). For the first speaker, the dataset contained approximately 50 hours of audio-book high quality data, which was aligned and phonetically transcribed. For AD, the subset contained only two hours of aligned data.

\begin{figure}[h]
  \centering
  \frame{\includegraphics[width=0.9\linewidth]{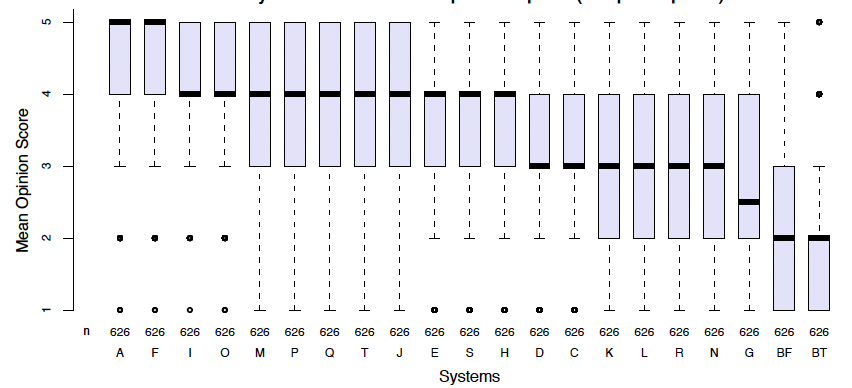}}
  \caption{Mean Opinion Scores for speaker NEB for native, non-speech expert evaluators - our system is identified by letter E and the human speaker is identified by letter A}
  \label{fig:results-neb-n-ne}
\end{figure}
\begin{figure}[h]
  \centering
  \frame{\includegraphics[width=0.9\linewidth]{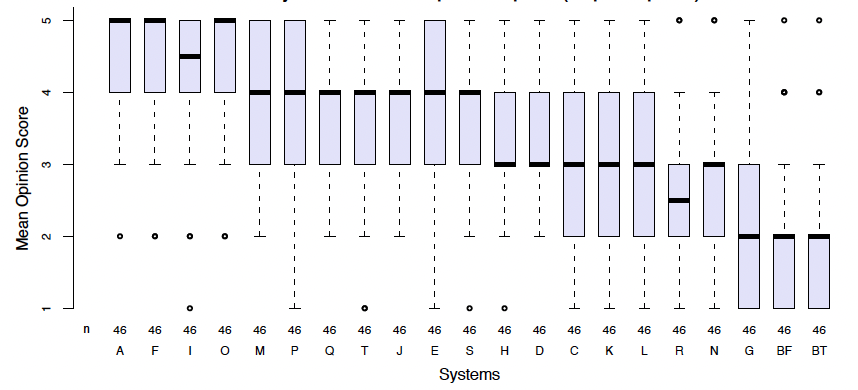}}
  \caption{Mean Opinion Scores for speaker NEB for native, speech expert evaluators - our system is identified by letter E and the human speaker is identified by letter A}
  \label{fig:results-neb-n-e}
\end{figure}
\begin{figure}[h]
  \centering
  \frame{\includegraphics[width=0.9\linewidth]{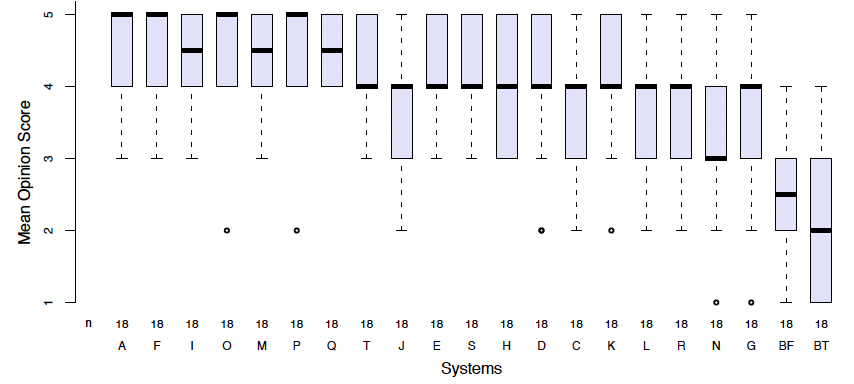}}
  \caption{Mean Opinion Scores for speaker NEB for non-native, non-speech expert evaluators - our system is identified by letter E and the human speaker is identified by letter A}
  \label{fig:results-neb-nn-ne}
\end{figure}
\begin{figure}[h]
  \centering
  \frame{\includegraphics[width=0.9\linewidth]{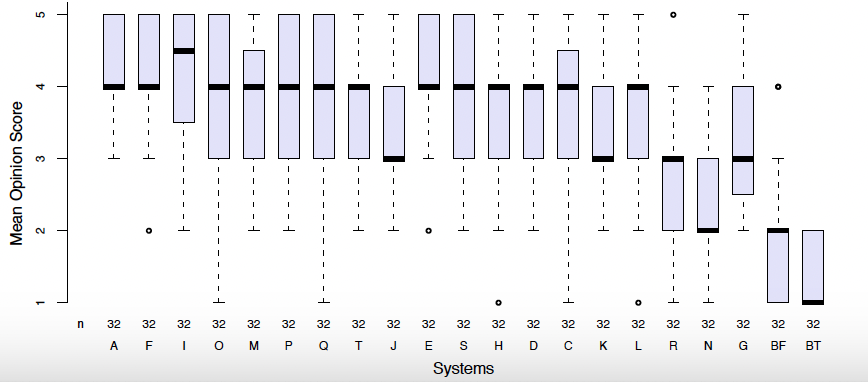}}
  \caption{Mean Opinion Scores for speaker NEB for non-native, speech expert evaluators - our system is identified by letter E and the human speaker is identified by letter A}
  \label{fig:results-neb-nn-e}
\end{figure}

\begin{figure}
  \centering
  \frame{\includegraphics[width=0.9\linewidth]{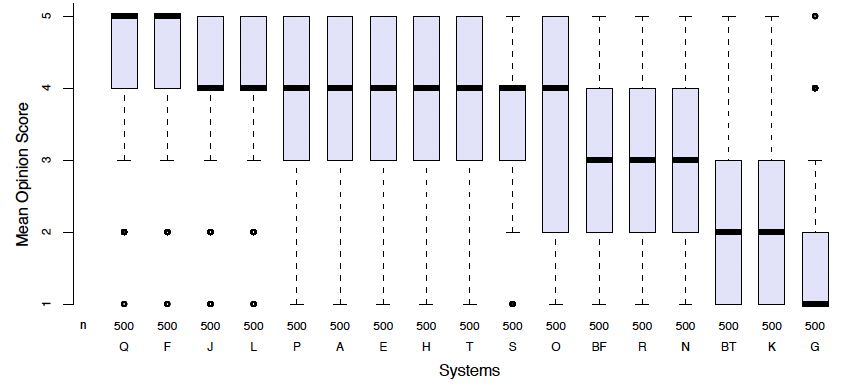}}
  \caption{Mean Opinion Scores for speaker AD for native, non-speech expert evaluators - our system is identified by letter E and the human speaker is identified by letter A}
  \label{fig:results-ad-n-ne}
\end{figure}

\begin{figure}
  \centering
  \frame{\includegraphics[width=0.9\linewidth]{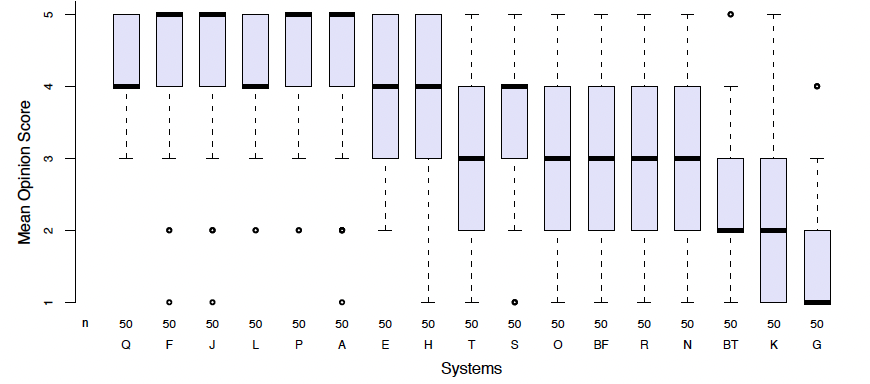}}
  \caption{Mean Opinion Scores for speaker AD for native, speech expert evaluators - our system is identified by letter E and the human speaker is identified by letter A}
  \label{fig:results-ad-n-e}
\end{figure}

\begin{figure}
  \centering
  \frame{\includegraphics[width=0.9\linewidth]{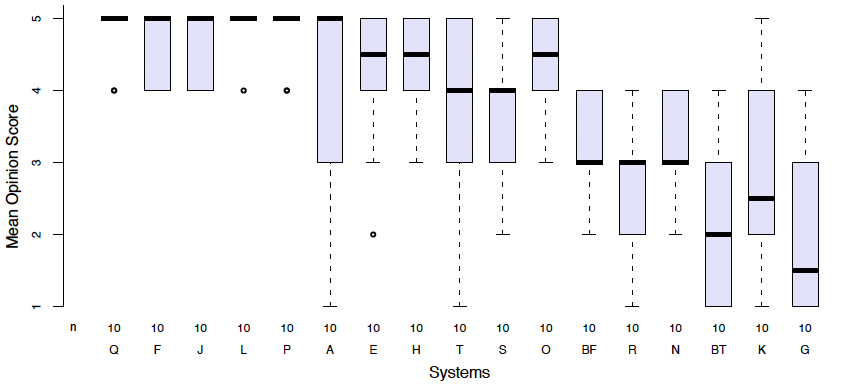}}
  \caption{Mean Opinion Scores for speaker AD for non-native, non-speech expert evaluators - our system is identified by letter E and the human speaker is identified by letter A}
  \label{fig:results-ad-nn-ne}
\end{figure}

\begin{figure}
  \centering
  \frame{\includegraphics[width=0.9\linewidth]{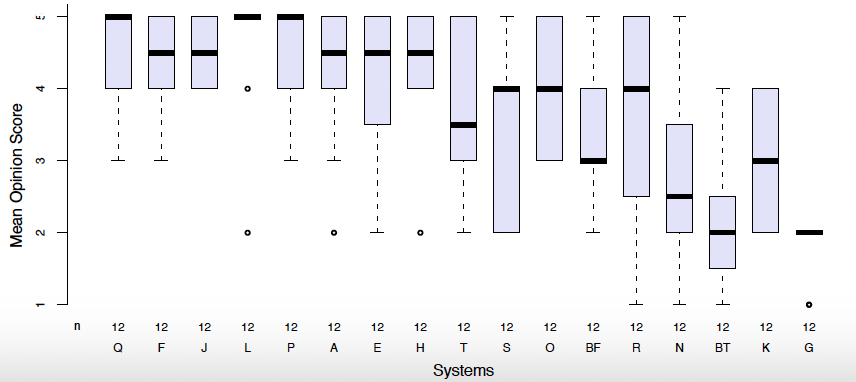}}
  \caption{Mean Opinion Scores for speaker AD for non-native, speech expert evaluators - our system is identified by letter E and the human speaker is identified by letter A}
  \label{fig:results-ad-nn-e}
\end{figure}

To compare our system with the other 20 that entered Blizzard Challenge we added fine-grained results as follows: (a) Figures \ref{fig:results-neb-n-e} and \ref{fig:results-ad-n-e} show the scores obtained from native speech experts for speakers NEB and AD respectively. (b) Figures \ref{fig:results-neb-n-ne} and \ref{fig:results-ad-n-ne} for native, non-speech experts, (c) Figures \ref{fig:results-neb-nn-e} and \ref{fig:results-ad-nn-e} for non-native speech experts 
and (d) Figures \ref {fig:results-neb-nn-ne} and \ref {fig:results-ad-nn-ne} for non-native and non-speech experts.


\begin{figure}
  \centering
  \frame{\includegraphics[width=0.9\linewidth]{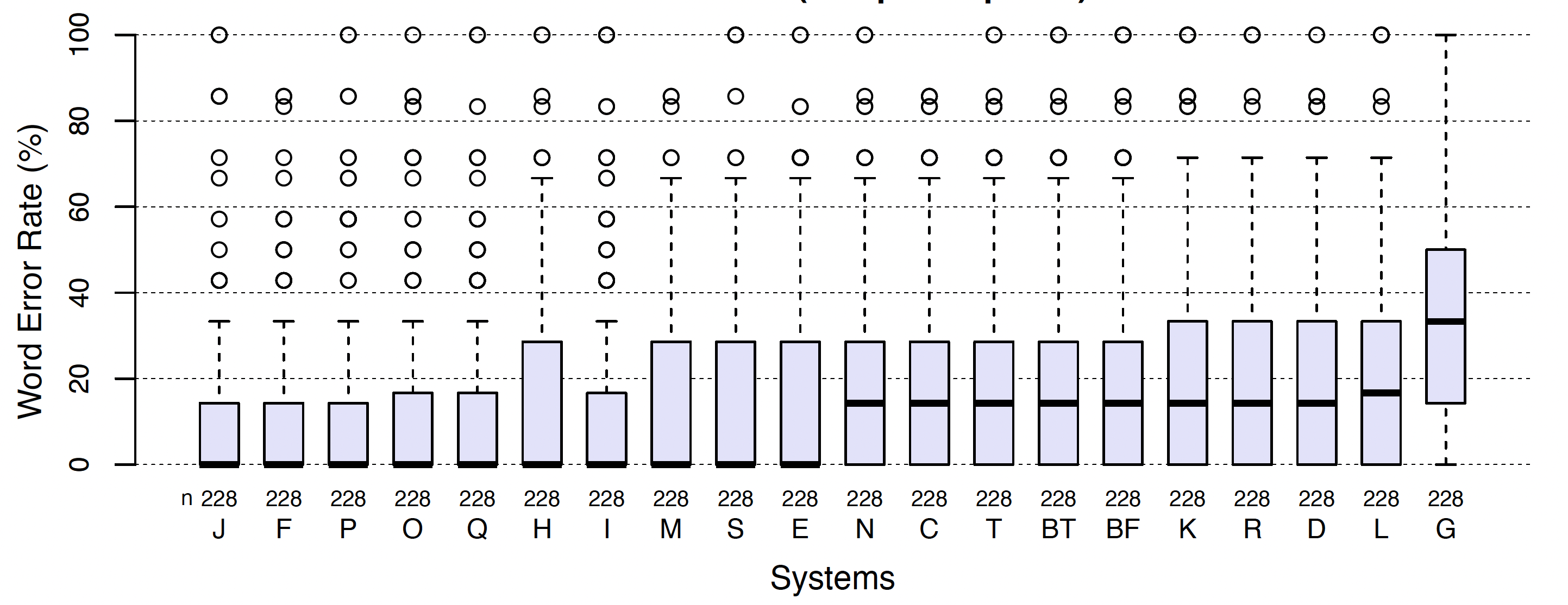}}
  \caption{Word error rates of all systems in the semantically unpredictable sentences test. The evaluation was run for speaker NEB}
  \label{fig:results-neb-wer}
\end{figure}

Table \ref{table:results} shows fine-grained Mean Opinion Scores (MOS) for both voices. It also takes into account if a user is French or not as well as his background (speech expert/non-expert). For a detailed comparison with the other systems in the challenge, we added fine-grained results, where our system is identified by the letter ``E''.

Overall, the system achieved good results, ranking 6 out of 20 systems and obtaining a MOS score above 4 in some sections. Interestingly, we achieved better scores from speech experts than from non-experts, which is not entirely expected, because speech experts are usually accustomed to spotting artefacts specific to speech synthesis. The scores from native french speakers were lower than those given by non-native speakers, which combined with the previous observation, might indicate that we had some issues with the phonetic transcription of the input. This is not something tested, but the intuition is that native speakers are able to spot mispronunciations far better than non-native speakers.

Additionally, Blizzard Challenge included an intelligibility test using semantically unpredictable sentences (SUS). In this test, the users are asked to transcribe utterances that followed a standard structure in terms of noun-verb-adjective order, but feature random words that belong to these classes. The purpose of having unpredictable input is to make the user rely on what he understands from listening to the SUSs, not infer the correct words based on his knowledge of the language. Our system obtained a word-error-rate (WER) of 0.162 with a standard deviation of 0.23, a results which is on-par with most of the top scoring system in the challenge. 

\section{Conclusions and future work}
\label{sect:conclusion}

We presented our entry for the Blizzard Challenge 2023: we provided an overview of our system architecture and provided details about individual components, and we included full training details with the hope that anyone interested will benefit from these details.

Furthermore, the system is fully open-sourced\footnote{https://github.com/tiberiu44/TTS-Cube} and we provide our pre-trained French models as well as a very simple API that allows users to synthesize their own text. We hope that our open contribution will enable result and model reproducibility and will encourage further research and development in the field of speech synthesis.

Simple API example: 
\begin{lstlisting}[language=Python]
from cube.api import TTSCube

cube=TTSCube.load("blizzard2023-hf")
speaker="neb"

audio=cube("Bonjour!", speaker=speaker)

IPython.display.Audio(data=audio, rate=24000)

\end{lstlisting}

The open-source system is compatible with TextGrid alignments and provides scripts for importing aligned data, training all models and building your own voice. Also, English models are included for download. The English model is based on the EN-HiFi corpus \cite{bakhturina2021hifi}.

\newpage
\clearpage

\bibliographystyle{IEEEtran}
\bibliography{mybib}

\begin{thebibliography}{10}
\providecommand{\url}[1]{#1}
\csname url@samestyle\endcsname
\providecommand{\newblock}{\relax}
\providecommand{\bibinfo}[2]{#2}
\providecommand{\BIBentrySTDinterwordspacing}{\spaceskip=0pt\relax}
\providecommand{\BIBentryALTinterwordstretchfactor}{4}
\providecommand{\BIBentryALTinterwordspacing}{\spaceskip=\fontdimen2\font plus
\BIBentryALTinterwordstretchfactor\fontdimen3\font minus
  \fontdimen4\font\relax}
\providecommand{\BIBforeignlanguage}[2]{{%
\expandafter\ifx\csname l@#1\endcsname\relax
\typeout{** WARNING: IEEEtran.bst: No hyphenation pattern has been}%
\typeout{** loaded for the language `#1'. Using the pattern for}%
\typeout{** the default language instead.}%
\else
\language=\csname l@#1\endcsname
\fi
#2}}
\providecommand{\BIBdecl}{\relax}
\BIBdecl

\bibitem{zen2007hmm}
H.~Zen, T.~Nose, J.~Yamagishi, S.~Sako, T.~Masuko, A.~W. Black, and K.~Tokuda,
  ``The hmm-based speech synthesis system (hts) version 2.0.'' \emph{SSW},
  vol.~6, pp. 294--299, 2007.

\bibitem{wang2017tacotron}
\BIBentryALTinterwordspacing
Y.~Wang, R.~Skerry-Ryan, D.~Stanton, Y.~Wu, R.~J. Weiss, N.~Jaitly, Z.~Yang,
  Y.~Xiao, Z.~Chen, S.~Bengio, Q.~Le, Y.~Agiomyrgiannakis, R.~Clark, and R.~A.
  Saurous, ``Tacotron: Towards end-to-end speech synthesis,'' in \emph{Proc.
  Interspeech 2017}, 2017, pp. 4006--4010. [Online]. Available:
  \url{http://dx.doi.org/10.21437/Interspeech.2017-1452}
\BIBentrySTDinterwordspacing

\bibitem{oord2016wavenet}
A.~{van den Oord}, S.~Dieleman, H.~Zen, K.~Simonyan, O.~Vinyals, A.~Graves,
  N.~Kalchbrenner, A.~Senior, and K.~Kavukcuoglu, ``{WaveNet: A Generative
  Model for Raw Audio},'' in \emph{Proc. 9th ISCA Workshop on Speech Synthesis
  Workshop (SSW 9)}, 2016, p. 125.

\bibitem{paine2016fast}
T.~L. Paine, P.~Khorrami, S.~Chang, Y.~Zhang, P.~Ramachandran, M.~A.
  Hasegawa-Johnson, and T.~S. Huang, ``Fast wavenet generation algorithm,''
  \emph{arXiv preprint arXiv:1611.09482}, 2016.

\bibitem{oord2018parallel}
A.~Oord, Y.~Li, I.~Babuschkin, K.~Simonyan, O.~Vinyals, K.~Kavukcuoglu,
  G.~Driessche, E.~Lockhart, L.~Cobo, F.~Stimberg \emph{et~al.}, ``Parallel
  wavenet: Fast high-fidelity speech synthesis,'' in \emph{International
  conference on machine learning}.\hskip 1em plus 0.5em minus 0.4em\relax PMLR,
  2018, pp. 3918--3926.

\bibitem{mehri2016samplernn}
\BIBentryALTinterwordspacing
S.~Mehri, K.~Kumar, I.~Gulrajani, R.~Kumar, S.~Jain, J.~Sotelo, A.~C.
  Courville, and Y.~Bengio, ``Samplernn: An unconditional end-to-end neural
  audio generation model,'' in \emph{5th International Conference on Learning
  Representations, {ICLR} 2017, Toulon, France, April 24-26, 2017, Conference
  Track Proceedings}.\hskip 1em plus 0.5em minus 0.4em\relax OpenReview.net,
  2017. [Online]. Available: \url{https://openreview.net/forum?id=SkxKPDv5xl}
\BIBentrySTDinterwordspacing

\bibitem{kalchbrenner2018efficient}
N.~Kalchbrenner, E.~Elsen, K.~Simonyan, S.~Noury, N.~Casagrande, E.~Lockhart,
  F.~Stimberg, A.~Oord, S.~Dieleman, and K.~Kavukcuoglu, ``Efficient neural
  audio synthesis,'' in \emph{International Conference on Machine
  Learning}.\hskip 1em plus 0.5em minus 0.4em\relax PMLR, 2018, pp. 2410--2419.

\bibitem{kim2018flowavenet}
\BIBentryALTinterwordspacing
S.~Kim, S.-G. Lee, J.~Song, J.~Kim, and S.~Yoon, ``{F}lo{W}ave{N}et : A
  generative flow for raw audio,'' in \emph{Proceedings of the 36th
  International Conference on Machine Learning}, ser. Proceedings of Machine
  Learning Research, K.~Chaudhuri and R.~Salakhutdinov, Eds., vol.~97.\hskip
  1em plus 0.5em minus 0.4em\relax PMLR, 09--15 Jun 2019, pp. 3370--3378.
  [Online]. Available: \url{https://proceedings.mlr.press/v97/kim19b.html}
\BIBentrySTDinterwordspacing

\bibitem{peng2020clarinet}
K.~Peng, W.~Ping, Z.~Song, and K.~Zhao, ``Non-autoregressive neural
  text-to-speech,'' in \emph{International conference on machine
  learning}.\hskip 1em plus 0.5em minus 0.4em\relax PMLR, 2020, pp. 7586--7598.

\bibitem{kumar2019melgan}
K.~Kumar, R.~Kumar, T.~De~Boissiere, L.~Gestin, W.~Z. Teoh, J.~Sotelo,
  A.~de~Br{\'e}bisson, Y.~Bengio, and A.~C. Courville, ``Melgan: Generative
  adversarial networks for conditional waveform synthesis,'' \emph{Advances in
  neural information processing systems}, vol.~32, 2019.

\bibitem{kong2020hifi}
J.~Kong, J.~Kim, and J.~Bae, ``Hifi-gan: Generative adversarial networks for
  efficient and high fidelity speech synthesis,'' \emph{Advances in Neural
  Information Processing Systems}, vol.~33, pp. 17\,022--17\,033, 2020.

\bibitem{devlin2018bert}
\BIBentryALTinterwordspacing
J.~Devlin, M.~Chang, K.~Lee, and K.~Toutanova, ``{BERT:} pre-training of deep
  bidirectional transformers for language understanding,'' in \emph{Proceedings
  of the 2019 Conference of the North American Chapter of the Association for
  Computational Linguistics: Human Language Technologies, {NAACL-HLT} 2019,
  Minneapolis, MN, USA, June 2-7, 2019, Volume 1 (Long and Short Papers)},
  J.~Burstein, C.~Doran, and T.~Solorio, Eds.\hskip 1em plus 0.5em minus
  0.4em\relax Association for Computational Linguistics, 2019, pp. 4171--4186.
  [Online]. Available: \url{https://doi.org/10.18653/v1/n19-1423}
\BIBentrySTDinterwordspacing

\bibitem{sotelo2017char2wav}
\BIBentryALTinterwordspacing
J.~M.~R. Sotelo, S.~Mehri, K.~Kumar, J.~F. Santos, K.~Kastner, A.~C. Courville,
  and Y.~Bengio, ``Char2wav: End-to-end speech synthesis,'' in
  \emph{International Conference on Learning Representations}, 2017. [Online].
  Available: \url{https://api.semanticscholar.org/CorpusID:30919574}
\BIBentrySTDinterwordspacing

\bibitem{lim2022jets}
D.~Lim, S.~Jung, and E.~Kim, ``Jets: Jointly training fastspeech2 and hifi-gan
  for end to end text to speech,'' in \emph{Proc. Interspeech 2022}, 09 2022,
  pp. 21--25.

\bibitem{liu2022delightfultts}
Y.~Liu, R.~Xue, L.~He, X.~Tan, and S.~Zhao, ``Delightfultts 2: End-to-end
  speech synthesis with adversarial vector-quantized auto-encoders,'' in
  \emph{Proc. Interspeech 2022}, 2022.

\bibitem{martin2019camembert}
\BIBentryALTinterwordspacing
L.~Martin, B.~Muller, P.~J. Ortiz~Su{\'a}rez, Y.~Dupont, L.~Romary,
  {\'E}.~de~la Clergerie, D.~Seddah, and B.~Sagot, ``{C}amem{BERT}: a tasty
  {F}rench language model,'' in \emph{Proceedings of the 58th Annual Meeting of
  the Association for Computational Linguistics}.\hskip 1em plus 0.5em minus
  0.4em\relax Online: Association for Computational Linguistics, Jul. 2020, pp.
  7203--7219. [Online]. Available:
  \url{https://aclanthology.org/2020.acl-main.645}
\BIBentrySTDinterwordspacing

\bibitem{rapt}
D.~Talkin, ``A robust algorithm for pitch tracking (rapt),'' in \emph{Speech
  Coding and Synthesis}, W.~B. Kleijn and K.~K. Paliwal, Eds.\hskip 1em plus
  0.5em minus 0.4em\relax Elsevier Science B.V., 1995, pp. 497--518.

\bibitem{kudo2018sentencepiece}
\BIBentryALTinterwordspacing
T.~Kudo and J.~Richardson, ``{S}entence{P}iece: A simple and language
  independent subword tokenizer and detokenizer for neural text processing,''
  in \emph{Proceedings of the 2018 Conference on Empirical Methods in Natural
  Language Processing: System Demonstrations}.\hskip 1em plus 0.5em minus
  0.4em\relax Brussels, Belgium: Association for Computational Linguistics,
  Nov. 2018, pp. 66--71. [Online]. Available:
  \url{https://aclanthology.org/D18-2012}
\BIBentrySTDinterwordspacing

\bibitem{gage1994bpeorig}
P.~Gage, ``A new algorithm for data compression,'' \emph{C Users Journal},
  vol.~12, no.~2, pp. 23--38, 1994.

\bibitem{sennrich2015bpe}
R.~Sennrich, B.~Haddow, and A.~Birch, ``Neural machine translation of rare
  words with subword units,'' in \emph{Proceedings of the 54th Annual Meeting
  of the Association for Computational Linguistics (Volume 1: Long
  Papers)}.\hskip 1em plus 0.5em minus 0.4em\relax Berlin, Germany: Association
  for Computational Linguistics, Aug. 2016, pp. 1715--1725.

\bibitem{bojanowski2017fasttext}
P.~Bojanowski, E.~Grave, A.~Joulin, and T.~Mikolov, ``Enriching word vectors
  with subword information,'' \emph{Transactions of the association for
  computational linguistics}, vol.~5, pp. 135--146, 2017.

\bibitem{blizzard-dataset}
\BIBentryALTinterwordspacing
G.~Bailly, O.~Perrotin, and M.~Lenglet, ``{Ressources for End-to-End French
  Text-to-Speech Blizzard challenge},'' Jan. 2023. [Online]. Available:
  \url{https://doi.org/10.5281/zenodo.7560290}
\BIBentrySTDinterwordspacing

\bibitem{bakhturina2021hifi}
\BIBentryALTinterwordspacing
E.~Bakhturina, V.~Lavrukhin, B.~Ginsburg, and Y.~Zhang, ``Hi-fi multi-speaker
  english tts dataset,'' in \emph{Interspeech}, 2021. [Online]. Available:
  \url{https://www.isca-speech.org/archive/pdfs/interspeech_2021/bakhturina21_interspeech.pdf}
\BIBentrySTDinterwordspacing

\end{thebibliography}

\end{document}